\RequirePackage{fix-cm}
\documentclass{svjour3}                     
\smartqed  
\usepackage{graphicx}

\usepackage{amssymb}
\usepackage{amsmath}
\usepackage[section]{placeins}
\usepackage[square,sort,comma,numbers]{natbib}
\journalname{Journal of Petroleum Exploration and Production Technology}
\begin{document}

\title{Gradient Boosting to Boost the Efficiency of Hydraulic Fracturing}

\author{Ivan Makhotin   \and
        Dmitry Koroteev \and
        Evgeny Burnaev
}


\institute{I. Makhotin \at
          Skolkovo Institute of Science and Technology \\
          \email{Ivan.Makhotin@skoltech.ru}
           \and
           D. Koroteev \at
           Skolkovo Institute of Science and Technology \\
           \email{D.Koroteev@skoltech.ru}
           \and
           E. Burnaev \at
           Skolkovo Institute of Science and Technology \\
           \email{E.Burnaev@skoltech.ru}  
}

\date{Received: date / Accepted: date}

\maketitle

\begin{abstract}
In this paper, we present a data-driven model for forecasting the production increase after hydraulic fracturing (HF). We use data from fracturing jobs performed at one of the Siberian oilfields. The data includes features, characterizing the jobs, and geological information. To predict an oil rate after the fracturing machine learning (ML) technique was applied. We compared the ML-based prediction to a prediction based on the experience of reservoir and production engineers responsible for the HF-job planning. We discuss the potential for further development of ML techniques for predicting changes in oil rate after HF.
\end{abstract}

\section{Introduction and Motivation}

This research is inspired by a practical issue of companies operating oilfields. The problem is related to a need for adequate forecasting of the efficiency of hydraulic fracturing jobs ~\cite{citation01, citation02, citation03, citation04}. An accurate prediction in terms of extra oil production allows performing reliable estimation of efficiency of investment in HF programs. Planning of HF programs typically includes two major parts. First is a selection of candidate wells for performing the HF jobs, and second is detailed planning of the jobs for each selected wellbore or a set of wellbores. The recent practice also includes planning the HF for newly drilled directional wells. It is quite common when we consider multistage HF as an essential part of well completion. This paper covers a novel approach for the second part of HF program planning.

It is well known, that uncertainty within a geological model of a hydrocarbon reservoir is a key source of risks at decision making for all the levels of field development planning workflows. Planning of a particular well stimulation job is not an exemption \cite{citation06}. This operation is typically based on a combination of physics-driven modeling of geomechanics and hydrodynamics of fracturing \cite{citation07, citation08} and further reservoir modeling with updated transport properties of near wellbore computational cells. The update of transport properties is typically driven by experience-based workflows formalized in an internal document of operating companies. There exist engineering approaches combining the reservoir modeling and fracture modeling (see \cite{citation09} for example) and the approaches already including machine learning for reservoir modeling \cite{citation10}.

\begin{figure}[t!]
  \centering
    \includegraphics[width=0.8\textwidth]{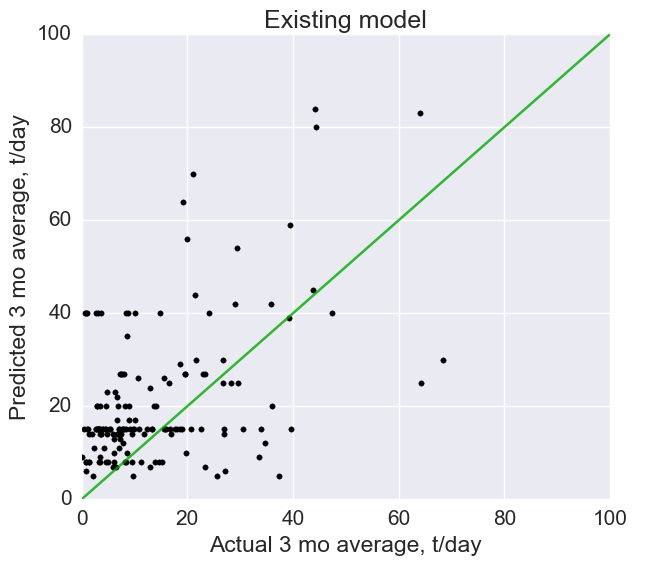}
   \caption{Prediction of extra production VS real data}
   \label{fig:existing_model_evaluation}
\end{figure}

A real example of utilization of the workflow described in the paragraph above is shown in Figure \ref{fig:existing_model_evaluation}. The scatter captures 270 wells from one of the West Siberian oilfields fractured from 2013 to 2016. The scatter contains wells with production history. One can immediately see that correlation between forecast and reality is very weak and errors in the forecast result in sufficient inaccuracy in a forecast of financial efficiency of the well stimulation program.  Also, one can see that predictions bellow 5 tons are thrown away despite indeed there are many wells which have actual average production rate below 5 tons. Moreover, one can spot a dense field of data points in the vicinity of 17 to 18 tons/day in the predicted values. This concentration reflects some bias of the experience-based forecasting which is likely related to a certain threshold of planned marginal profit out of the fractional job.

In many cases machine learning allow handling the forecasting problems for cases when the accuracy of physics-driven or empirical models is limited by uncertainties of their input parameters and can provide a fast approximation, or the so-called surrogate models, to estimate selected properties based on the results of real measurements (for details, see \citep{GTApprox2016}). Surrogate models are a well-known way of solving various industrial engineering problems including oil industry \citep{grihon2013surrogate,SMspacecraft2016, Grihon2014, GrihonFactorial2014,drilling2019,oil2019,core2019}.

This particular study shows the potential of the machine learning in a case when we do not know the exact values characterizing the geological and physical state of the formation targeted for a hydraulic fracturing but have a history of fracturing jobs conducted in the wells of the same oilfield. 

\section{Methods}

The main purpose was to obtain an approximation $\hat{F}(x)$ of the function $F(x)$ mapping $x$ to $y$, using the training set $\{x_i, y_i\}_{i=1}^{n}$. Here $x \in \mathbb{R}^d$ is a vector of $d$ parameters which describes fracturing job and geology of a well, $y$ is an oil rate per day averaged for three months after corresponding HF. Having the loss function $L(\hat{F}(x), y)$ the goal is to minimize estimation of the expected value of the loss over the joint distribution of all $(x, y)$ values. In practice we use an empirical distribution, i.e., we would like to construct
\[
\hat{F} = \underset{F}{\mathrm{arg}\min}\, \hat{\mathbb{E}}_{x,y} L(F(x), y) = \underset{F}{\mathrm{arg}\min}\, \frac{1}{n} \sum_{i=1}^{n} L(F(x_i), y_i).
\]

Gradient boosting \cite{citation11} is one of the machine learning algorithms for regression, and it proved itself to be robust, reliable and sufficiently accurate for engineering applications. Gradient boosting combines $M$ weak estimators $h_m$ into a single strong estimator $F_M$ in an iterative way:
\begin{align*}
F_M(x) &= \sum_{m=1}^{M} b_m h_m(x, a_m),\ b_m \in \mathbb{R},\ a_m \in A,\\
Q_M &= \sum_{i=1}^{n} L(y_i, F_{M-1}(x_i) + b_M h_M(x_i, a_M)) \to\min_{b_M,a_M}.
\end{align*}

The gradient boosting algorithm improves on $F_{M-1}$ by constructing new weak estimator $h_M$ and adding it to the general model with appropriate coefficient $b_M$. The idea is to apply gradient descent in a functional space of estimators and fit a new weak estimator to the anti-gradient of the loss function:
\begin{align*}
&[\nabla Q^i_M]_{i=1}^n = \Bigg[\frac{\partial Q_M}{\partial F_{M-1}(x_i)}\Bigg]_{i=1}^{n},\,\,a_M = \underset{a \in A}{\mathrm{arg}\min} \sum_{i=1}^{n} ((-\nabla Q^i_M)-h_M(x_i, a))^2,\\
&\mbox{Gradient step:}\,\,F_M(x) = F_{M-1}(x) + b h_M(x, a_M),\ b \in \mathbb{R}.
\end{align*}

Finally we find gradient step size $b_M$ with a simple one-dimensional optimization 
\[b_M = \underset{b \in \mathbb{R}}{\mathrm{arg}\min} \sum_{i=1}^{n} L(y_i,F_{M-1}(x_i) + bh_M(x_i, a_M)).
\]

Trees of small depth are used as weak estimators. We tuned the number of iterations $M$ and maximal tree depth using cross-validation. 
Gradient boosting over decision trees is known to produce relatively high accuracy forecasts while operating with datasets having even a significant amount of missing data, which is the case for this study. 

We divided input parameters for our machine learning model into five groups:
\begin{itemize}
    \item[1.] General Information: Well number, fracturing job date, time when treatment started, zone, contractor (company), supervisor name (from contractor side), supervisor name (from client side), fracturing status (initial fracturing or refracturing), well status (new drill or old well), completion type (cemented or open hole), number of stages,
    \item[2.] Job Parameters: flow rate, pad volume, total volume of gelled fluid pumped, maximal concentration of proppant, proppants’ (at 1st to 4th stage of frac job) manufacturers, proppants’ mesh sizes and volumes, data-frac pumped (Yes/No), fracture closure gradient (kPa/m), instantaneous shut-in pressure gradient, fracture net pressure, maximal wellhead treating pressure, average wellhead treating pressure, actual - planned flush (0 if equal, 1 if positive, -1 if negative), screen-out (Yes/No),
    \item[3.] Fluid Parameters: gel type (Guar, HPG, …), gel loading (kg/m3), types and amounts of breakers, types and amounts of X-Linkers used at different stages.
    \item[4.] Calculated HF parameters: Estimated fracture height (m), Estimated fracture length (m), Estimated fracture width (mm).
    \item[5.] Geological data: Clay factor (relative units), porosity (\%), thickness of gas saturated part (m), reservoir thickness (m), thickness of the target interval (m), length of horizontal part of the wellbore (m), oil saturation (\%), thickness of oil-saturated interval (m), sand content (\%), permeability (mD), reservoir compartmentalization index, reservoir bottom depth (m).
\end{itemize}

The target value is oil rate per day averaged for three months after corresponding HF. One can see the histograms of the selected parameters in the appendix in Fig. \ref{fig:distributions}.

The dataset contains many categorical parameters. To transform these parameters in a numeric form, we used one-hot-encoding approach. This approach allows transforming a categorical parameter to a boolean vector. Each position in that vector is associated with the corresponding unique category. The categorical parameter can be encoded as such vector with one at corresponding position and zeros in the others. After that, we obtain a final input vector $x$ as a concatenation of a vector with all available numeric parameters and all vectors which encode categorical features. This approach makes sense if categorical parameter has a small number of unique values and so $d<n$, otherwise we can get dimensionality explosion.

All samples are split into the training set (80\%) and the test set (20\%). We fit the model using samples from the training set and calculate an average error using the test set. For more accurate generalization ability estimation, one can randomly divide samples into train/test subsets several times and average the error over these divisions as well. In this paper we provide the average error estimated using 50 random splits. As the error measure we considered MAE and Pearson correlation coefficient. Here
\[
MAE = \frac{1}{n} \sum_{i=1}^{n}\lvert\hat{y}_i - y_i\rvert.
\]

\section{Results}

\begin{figure}[t!]
  \centering
    \includegraphics[width=1.\textwidth]{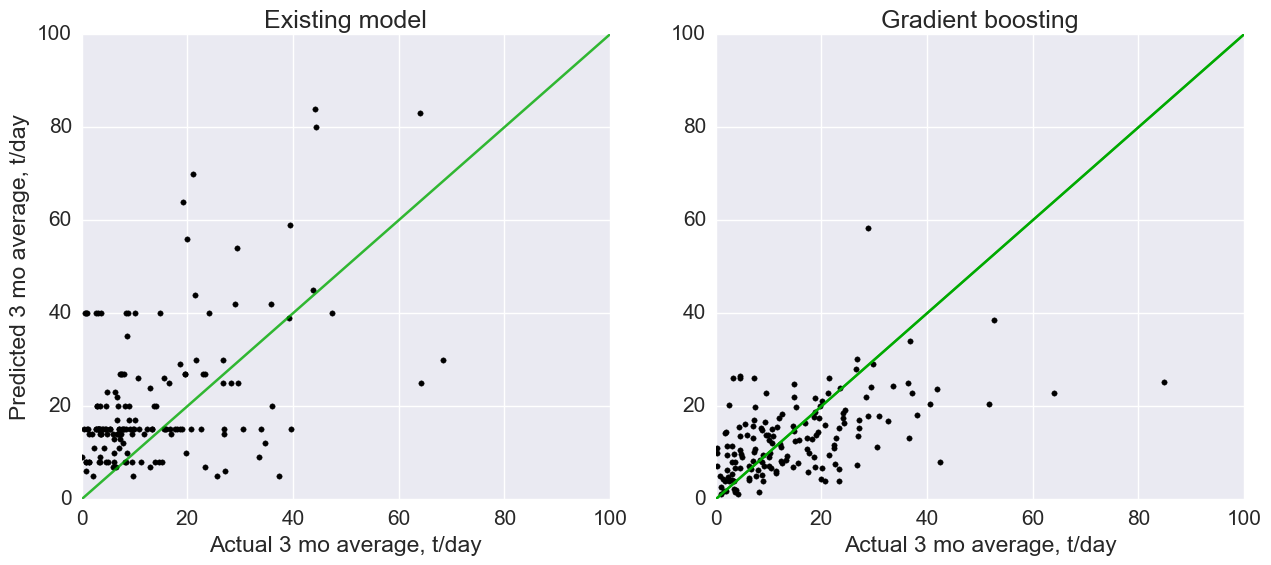}
   \caption{Existing model vs gradient boosting model (20\% test)}
   \label{fig:gb_vs_existing}
\end{figure}

\begin{table}[t!] 
\centering 
\caption{Comparison of models performance. Metrics on 20\% test averaged by 50 random train\textbackslash test splits. For existing model, metrics were calculated over all known predictions} 
\label{tab:comparison}
\smallskip\noindent
\begin{tabular}{|l|l|l|l|l|l|l|l|} 
\hline 
   & Existing model & Gradient Boosting \\ \hline 
MAE & 12.23 & 9.68\\ \hline 
Pearson corr. coeff. & 0.47 & 0.63 \\ \hline 

\end{tabular}
\end{table}

Predictions of the existing empirical model for 20\% of the cases are depicted in the left plot (see Fig. \ref{fig:gb_vs_existing}). In the right plot, we depicted predictions of the gradient boosting model also for 20\% of the samples; other 80\% were used for model training. One can see that the machine learning model has much higher forecasting ability. This can be seen by comparing the average error and the correlation of predictions with the existing model (see Table \ref{tab:comparison}).

\begin{figure}[t!]
  \centering
    \includegraphics[width=1.\textwidth]{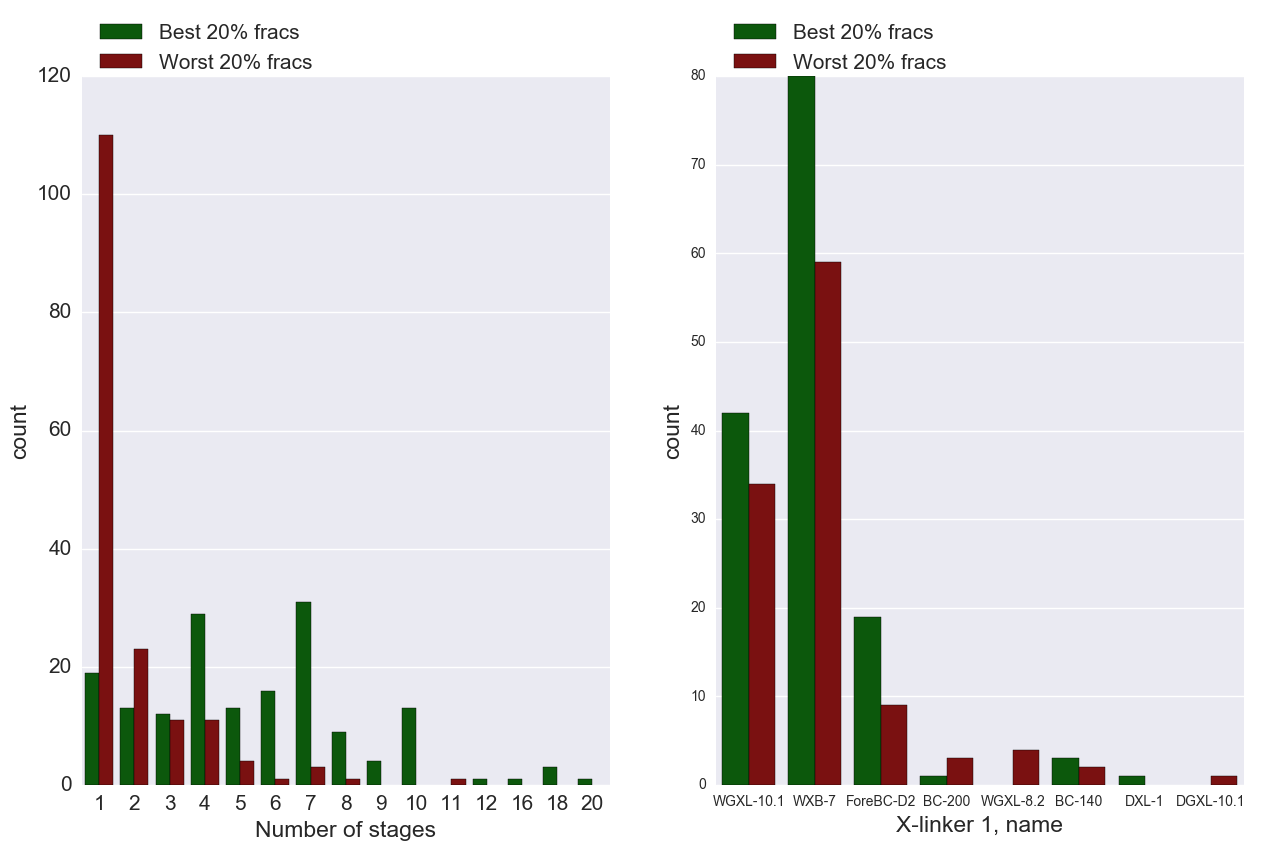}
   \caption{Number of stages and X-linker for best/worst}
   \label{fig:stat1}
\end{figure}

\begin{figure}[t!]
  \centering
    \includegraphics[width=1.\textwidth]{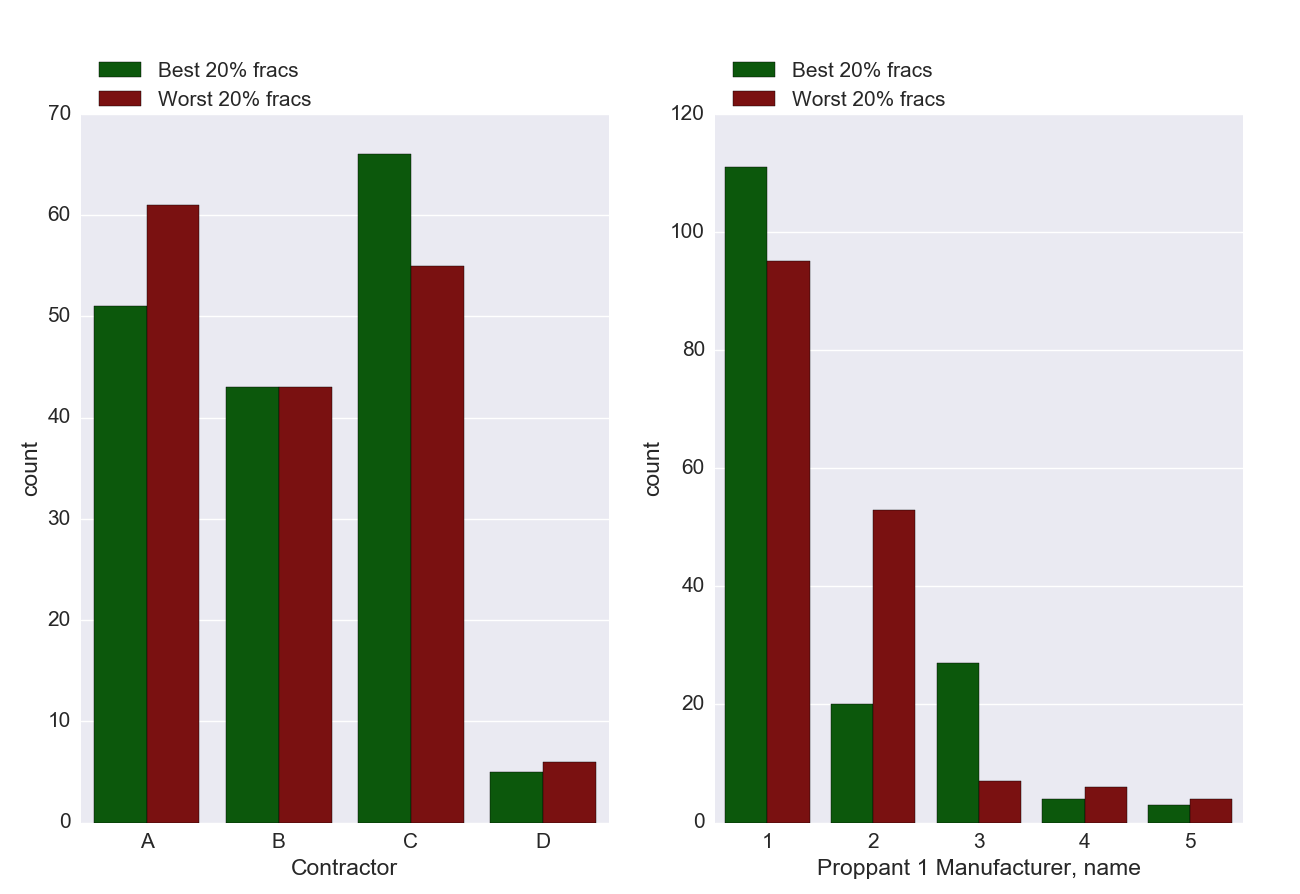}
   \caption{Contractor and proppant manufacturer for best/worst}
   \label{fig:stat2}
\end{figure}

Purely statistical observations can give some valuable insights about the data. For example, in Fig. \ref{fig:stat1} and Fig. \ref{fig:stat2} one can see the distribution of categorical features for either best (in terms of extra oil after the fracturing job) 20\% fracs or worst 20\% fracs. In Fig. \ref{fig:stat1} left plot shows that fracs with many stages are more successful than others. An interesting observation is that seven stage fracturing looks optimal in terms of gaining a maximal amount of extra oil and with ten stage fracturing one can expect no poorly producing wells. Right plot shows, for example, that with WGXL-8.2 and DGXL-10.1 X-linker there are only the worst fracs and perhaps one should exclude them. 

In Fig. \ref{fig:stat2} left plot shows that contractor C provides the most successful fracturing services and contractor A is the worst one in terms of relative quality. But the actual differences are relatively small and can be interpreted that success does not strongly depend on the contractor. Also in the right plot, one can see which proppant manufacturer is better. The names of proppant manufacturers and contractors were changed for reasons of confidentiality.

\section{Conclusions and Discussion}

One can spot (see Fig. \ref{fig:gb_vs_existing}), that the experience-based model tends to overestimate production rates over the whole range of the actual flow rates, while the gradient boosting generates underestimates at the high flow rates. The gradient boosting behaves like this because it targets exactly minimizing the average error and there are very few samples of high flow rate cases within the training and validation sets. From the economical standpoint, the authors believe that such performance of the gradient boosting based model is somewhat safe as it allows managing the expectations of outcomes of an HF job in a conservative manner.

In overall, the paper demonstrates usability and a very high potential of machine learning technologies as a tool for prediction of hydraulic fracturing efficiency. This is just a first effort of bringing the modern big data techniques to the well stimulation optimization. Authors believe that this is just an initial step in these directions. There are multiple ways of improvement of the existing models. They include an accurate assessment of prediction error e.g. using nonparametric confidence measures \cite{VovkConformal2014,ConformalKRR2016}, precise selection of objective function for optimization of the algorithms, comparing different training routines, applying smart algorithms for filling the gaps in the initial data and assessing the data quality, performing feature engineering, detailed comparison with other machine learning methods. 

It is rather obvious that further improvement of data-driven forecasting algorithms and data collection systems will make machine learning a true game-changer for the upstream technologies resulting in sufficient optimization of the technological and economical side of the processes generating real data.

\section*{Appendix}
\begin{figure}[t!]
  \centering
    \includegraphics[width=0.83\textwidth]{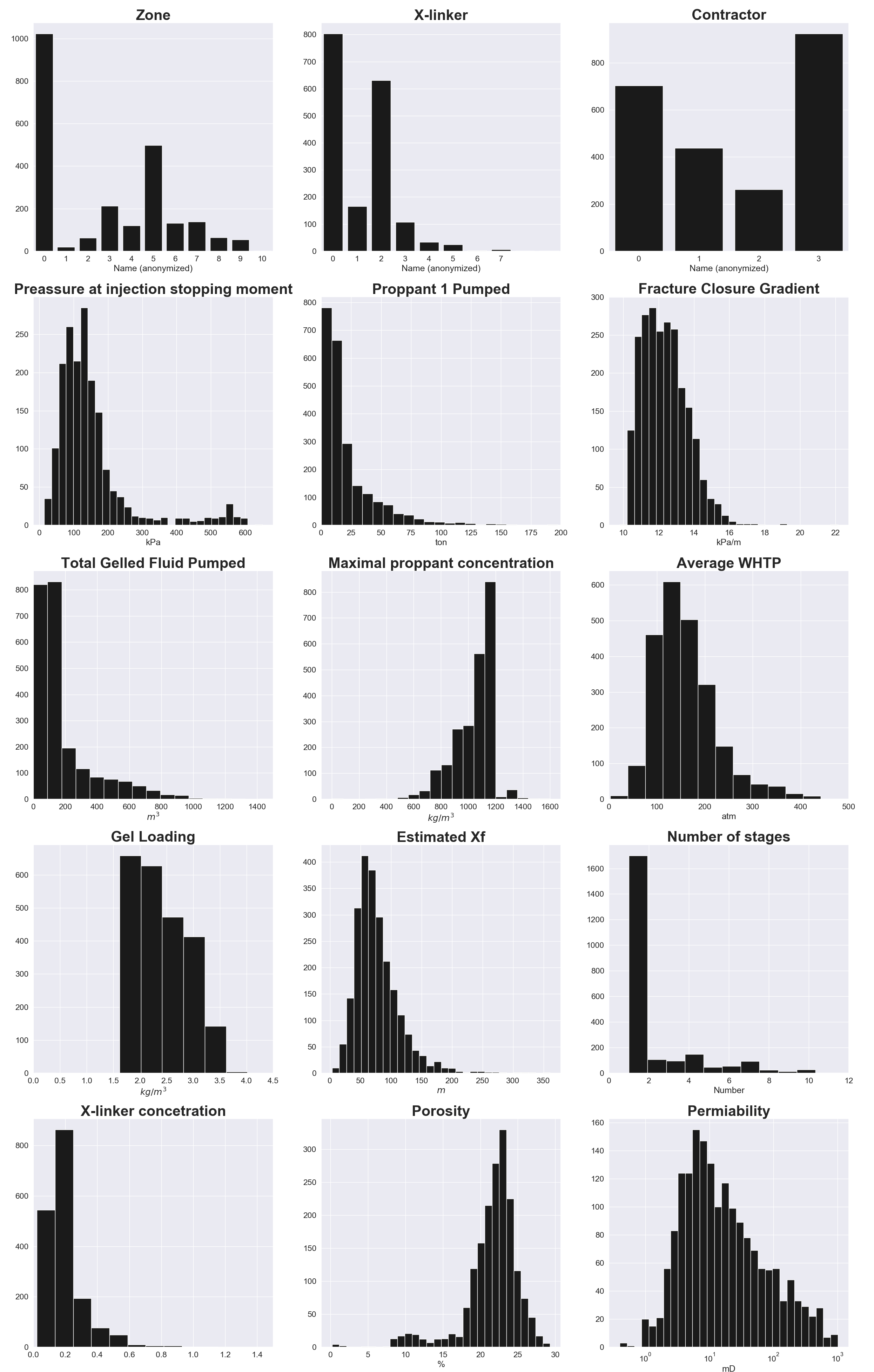}
   \caption{Distributions of the selected parameters in the initial dataset
   }
   \label{fig:distributions}
\end{figure}

There are histograms for several selected variables in Fig. \ref{fig:distributions}. One can see that in historical data, the HFs were made for several zones and with different contractors. The most of hydraulic fracturing jobs are single-stage. The remaining parameters are distributed without any significant anomalies.

\bibliographystyle{spmpsci}      
\bibliography{bibfile}   

\end{document}